\documentclass[lettersize,journal]{IEEEtran}
\usepackage{amsmath,amsfonts}
\usepackage[ruled,linesnumbered]{algorithm2e} 
\usepackage{array}
\usepackage[caption=false,font=normalsize,labelfont=sf,textfont=sf]{subfig}
\usepackage{textcomp}
\usepackage{stfloats}
\usepackage{url}
\usepackage{amssymb}
\usepackage{booktabs}
\usepackage{foreign}
\usepackage{verbatim}
\usepackage{booktabs}
\usepackage{multirow}
\usepackage{graphicx}
\usepackage{graphicx}
\usepackage{xcolor}

\usepackage{cite}
\begin{document}

\title{Semantic Mismatch and Perceptual Degradation: A New Perspective on Image Editing Immunity}

\author{
Shuai Dong~\IEEEmembership{,}
\and
Jie Zhang,~\IEEEmembership{Member,~IEEE,}
\and
Guoying Zhao~\IEEEmembership{Fellow,~IEEE,}
\and
Shiguang Shan~\IEEEmembership{Fellow,~IEEE,}
\and
Xilin Chen~\IEEEmembership{Fellow,~IEEE,}

\thanks{Jie Zhang, Shiguang Shan and Xilin Chen are with the State Key Laboratory of AI Safety, Institute of Computing Technology, Chinese Academy of Sciences (CAS), Beijing 100190, China, and also with the University of China Academy of Sciences, Beijing 100049, China (e-mail: zhangjie@ict.ac.cn; sgshan@ict.ac.cn; xlchen@ict.ac.cn).}
\thanks{Guoying Zhao is with the Center for Machine Vision and Signal Analysis,
University of Oulu, 90014 Oulu, Finland (e-mail: guoying.zhao@oulu.fi).}
\thanks{Shuai Dong is with the School of Computer Science, China University of Geosciences, Wuhan 430074, China (e-mail: dongshuai\_iu@163.com).}
}
\markboth{Journal of \LaTeX\ Class Files,~Vol.~14, No.~8, August~2021}%
{Shell \MakeLowercase{\textit{et al.}}: A Sample Article Using IEEEtran.cls for IEEE Journals}

\IEEEpubid{0000--0000/00\$00.00~\copyright~2021 IEEE}

\maketitle

\begin{abstract}
Text-guided image editing via diffusion models, while powerful, raises significant concerns about misuse, motivating efforts to immunize images against unauthorized edits using imperceptible perturbations. Prevailing metrics for evaluating immunization success typically rely on measuring the visual dissimilarity between the output generated from a protected image and a reference output generated from the unprotected original. This approach fundamentally overlooks the core requirement of image immunization, which is to disrupt semantic alignment with attacker intent, regardless of deviation from any specific output. We argue that immunization success should instead be defined by the edited output either semantically mismatching the prompt or suffering substantial perceptual degradations, both of which thwart malicious intent. To operationalize this principle, we propose Synergistic Intermediate Feature Manipulation (SIFM), a method that strategically perturbs intermediate diffusion features through dual synergistic objectives: (1) maximizing feature divergence from the original edit trajectory to disrupt semantic alignment with the expected edit, and (2) minimizing feature norms to induce perceptual degradations. Furthermore, we introduce the Immunization Success Rate (ISR), a novel metric designed to rigorously quantify true immunization efficacy for the first time. ISR quantifies the proportion of edits where immunization induces either semantic failure relative to the prompt or significant perceptual degradations, assessed via Multimodal Large Language Models (MLLMs). Extensive experiments show our SIFM achieves the state-of-the-art performance for safeguarding visual content against malicious diffusion-based manipulation.
\end{abstract}

\begin{IEEEkeywords}
Diffusion Models, Image Editing, Image Immunization.
\end{IEEEkeywords}

\section{Introduction}
\IEEEPARstart{R}{ecent} breakthroughs in diffusion models (DMs) have revolutionized text-to-image synthesis and manipulation ~\cite{ho2020denoising,Song2019GenerativeMB,Dhariwal2021DiffusionMB,Nichol2021ImprovedDD,Watson2022LearningFS,Rombach2021HighResolutionIS,Croitoru2022DiffusionMI,11267093,10374263}, enabling high-fidelity generation and fine-grained editing through natural language guidance ~\cite{Avrahami2021BlendedDF,Couairon2022DiffEditDS,Meng2021SDEditGI,Mokady2022NulltextIF,Wallace2022EDICTED,Wang2022ImagenEA,Xie2022SmartBrushTA,Xu2023OpenVocabularyPS,Zhu2023MovieFactoryAM,zhao2024ultraeditinstructionbasedfinegrainedimage,10976616,10313073}. While these capabilities unlock transformative creative tools, they also introduce severe ethical risks. Malicious actors could exploit DMs to generate deepfakes or forged content for disinformation campaigns, privacy violations, or manipulation of public discourse~\cite{Pei2024DeepfakeGA}. Such threats erode trust in digital media and risk destabilizing socio-political systems, making the development of robust safeguards against harmful edits a critical priority. One promising defense paradigm is image immunization, which addresses this challenge by embedding imperceptible perturbations into images to proactively disrupt unauthorized edits ~\cite{Aneja2021TAFIMTA,Ruiz2020DisruptingDA,Ruiz2023PracticalDO,Yeh2021AttackAT,Salman2023RaisingTC,Chen2024EditShieldPU,Zheng2023TargetedAI,Xue2023TowardEP,Liang2023MistTI}.

\begin{figure*}[t!]
  \centering
  \includegraphics[width=1\textwidth]{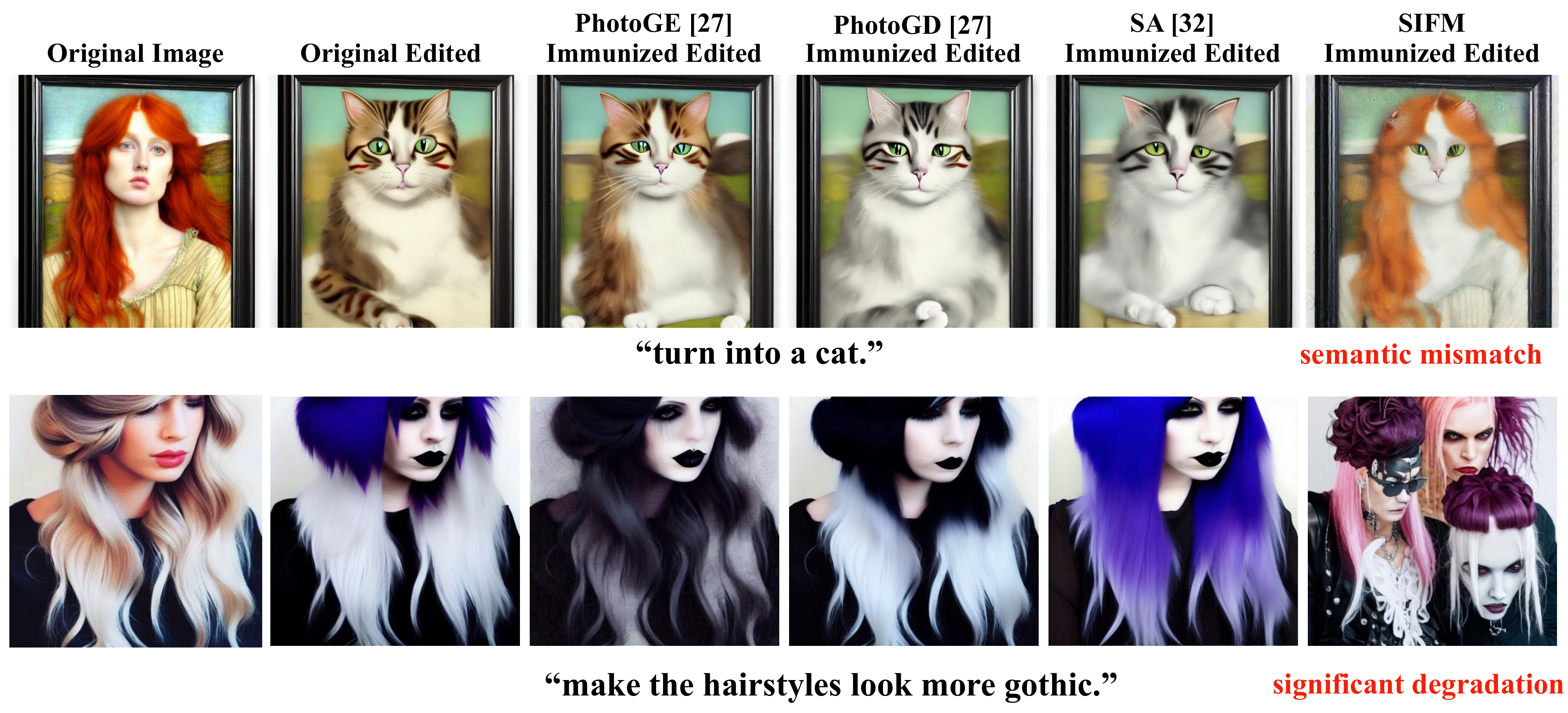} % 替换为您的图片文件名
  % \vspace{-10pt}
  \caption{The image editing immunization effects across various methods. Although methods like PhotoGE~\cite{Salman2023RaisingTC}, PhotoGD~\cite{Salman2023RaisingTC}, and SA~\cite{Lo_2024_CVPR} produce immunized edits that differ visually from the original edited results, their outputs still adhere to the edit instructions (i.e., immunization failure). In contrast, our proposed SIFM successfully achieves image immunization. Notably, ``Immunized Edited'' refers to the edited output of the immunized image.}
  \label{fig:example}
\end{figure*}

Despite notable advancements in mitigating malicious edits~\cite{Salman2023RaisingTC,Lo_2024_CVPR,Chen2024EditShieldPU,Zheng2023TargetedAI,Xue2023TowardEP,Liang2023MistTI}, the prevailing standard for defining immunization success remains superficial, predominantly relying on visual dissimilarity between the edited output of the immunized image and the specific edited output of the unprotected original. This approach is fundamentally limited because such a reference edit represents merely one possible outcome among a spectrum of valid edits for a given prompt, especially given the inherent variability of diffusion models. Consequently, deviation from such a non-unique reference does not inherently signify successful immunization. Critically, these evaluations overlook the core requirement that effective protection must disrupt semantic alignment with the attacker's intent, irrespective of comparison to any specific reference instance. For example, with the prompt ``make the hairstyles look more gothic'' as shown in Fig.~\ref{fig:example}, various ``gothic'' interpretations can be semantically valid yet look very different from the edited original. Therefore, significant visual deviation alone does not mean a malicious edit has been prevented. This inadequacy of current standards to genuinely assess immunization raises a critical question: What constitutes an accurate standard for immunization success?

\IEEEpubidadjcol
To address the question, we redefine successful image immunization through an adversarial lens: the success criterion for immunization is the neutralization of the attacker's intent, not the divergence of the output from its original form. Specifically, we argue that immunization is successful if the edited output either \textbf{1) fails to align semantically with the intent of the guiding text prompt} or \textbf{2) suffers from substantial perceptual degradations}, such as artifacts or quality degradation, rendering it unusable for malicious purposes. This dual-axis definition shifts focus from superficial visual dissimilarity to practically consequential outcomes, ensuring immunization effectiveness aligns with adversarial intent.

Guided by our redefined immunization success objectives, we propose Synergistic Intermediate Feature Manipulation (SIFM), a defense mechanism that strategically disrupts edits through dual interventions in diffusion model features. Specifically, building on the observation that deeper diffusion layers encode semantically rich~\cite{Naseer2021IntriguingPO,Raghu2021DoVT}, SIFM maximizes the distance between the intermediate feature trajectory of immunized and non-immunized edits. By perturbing features at critical semantic bottlenecks, this objective directly induces semantic mismatch, ensuring the edited output diverges fundamentally from the attacker’s intent. Simultaneously, SIFM minimizes the Frobenius norm of targeted intermediate features. This destabilizes the generation process, introducing significant degradations that degrade output usability. By synergistically coupling these objectives, SIFM directly enforces semantic misalignment with the edit prompt and perceptual degradations, thereby invalidating the edit’s plausibility for immunization success.

Having redefined immunization success and introduced SIFM to enforce its dual criteria, another crucial question remains: How can we reliably assess whether an edited immunized outputs achieve semantic mismatch or perceptual degradations? As previously established, evaluation paradigms such as those employing PSNR or LPIPS~\cite{Zhang2018TheUE} remain superficial. Their predominant reliance on visual dissimilarity between an edited immunized image and a single, non-unique reference edit of the original renders them incapable of accurately assessing true immunization success. This is because they fail to consider two key aspects: the existence of multiple valid edit outcomes and, more importantly, the semantic alignment between the result and the prompt. This limitation necessitates a move towards evaluation methods capable of more sophisticated semantic reasoning.

We address this gap by proposing the Immunization Success Rate (ISR), a novel evaluation metric grounded in multimodal reasoning and degradation analysis. We leverage Multimodal Large Language Models (MLLMs) to qualitatively assess if an edited immunized image either (1) significantly deviates semantically from the prompt or (2) exhibits perceptual degradations, rendering it implausible. By prompting MLLMs to analyze semantic alignment and perceptual integrity, we bypass the limitations of former evaluation metrics. Each edited instance of an immunized image is adjudicated as a successful defense only if it satisfies either criterion above. Aggregating these binary judgments across a dataset yields the Immunization Success Rate (ISR), a quantitative metric reflecting the proportion of edits neutralized by immunization.

In summary, our main contributions are:

\begin{itemize}
    \item We critique prevailing evaluation methods anchored in pairwise image evaluation metrics (e.g., PSNR and LPIPS) and propose, for the first time, a semantically grounded definition of immunization success: an edit is neutralized if it either 1) fundamentally misaligns with the edit prompt’s intent or 2) incurs significant perceptual degradations, rendering the output implausible.
    \item SIFM, a novel immunization method that perturbs intermediate diffusion features is introduced to induce both semantic deviation from the prompt and significant perceptual degradations.
    \item We propose the Immunization Success Rate (ISR), a new metric leveraging Multimodal LLMs to better quantify immunization success.
\end{itemize}

\section{Related work}
In this section, we review the literature pertinent to our proposed framework, categorizing the discussion into two primary domains. First, we survey Image Immunization Against Generative Models to situate our SIFM method within existing image immunizations and to critique the prevailing evaluation standards for their dependence on superficial visual distance. Second, we review multimodal models for evaluation to ground our proposed Immunization Success Rate (ISR). These models offer the advanced semantic reasoning required to overcome the shortcomings of pixel-level metrics, enabling a more accurate assessment of immunization.

\subsection{Image Immunization Against Generative Models.}
The proliferation of diffusion models enabling facile text-guided image editing~\cite{Rombach2021HighResolutionIS, Brooks2022InstructPix2PixLT} underscores the urgent need for robust defenses against misuse~\cite{Salman2023RaisingTC, Chen2024EditShieldPU}. Early image immunization efforts include PhotoGE and PhotoGD by Salman \textit{et al.}~\cite{Salman2023RaisingTC}, which inject imperceptible perturbations to disrupt diffusion models by modifying VAE-encoded latents or the entire sampling process to degrade outputs. Lo \textit{et al.}~\cite{Lo_2024_CVPR} introduces Semantic Attack (SA) for targeted disruption through selective manipulation of text-relevant cross-attention regions. Further approaches include EditShield (ED) by Chen \textit{et al.}~\cite{Chen2024EditShieldPU}, target instruction-guided edits by shifting latent representations to induce mismatched subjects. Score Distillation Sampling (SDS)~\cite{Xue2023TowardEP} utilizes score distillation sampling for efficient Latent Diffusion Model encoder-targeting perturbations against mimicry. Liang and Wu propose MIST~\cite{Liang2023MistTI} with textural losses for robust protection against image imitation and unauthorized customization, while Attacking with Consistent score-function Errors (ACE) by Zheng et al.~\cite{Zheng2023TargetedAI} employs targeted attacks creating consistent score-function errors to poison diffusion models against customization. Despite varied adversarial strategies, prevailing immunization evaluations use distance metrics like PSNR or LPIPS~\cite{Zhang2018TheUE}. These metrics compare the edited output of the immunized image against the corresponding edited output of the unprotected original, a pixel-level focus that critically overlooks semantic prompt alignment. This insufficiency in assessing true immunization success motivates our development of semantically-aware evaluations.

\subsection{Multimodal Models for Evaluation Tasks.}
The inherent limitations of pairwise image similarity metrics (PSNR and LPIPS~\cite{Zhang2018TheUE}) necessitate a paradigm shift toward more sophisticated evaluation tools. Specifically, their inability to capture semantic context or intended visual changes renders them inadequate for measuring true immunization success. Recent advancements in Multimodal Large Language Models (MLLMs)~\cite{Lu2019ViLBERTPT,Wu2024DeepSeekVL2MV,Tan2019LXMERTLC,Alayrac2022FlamingoAV,Li2024LLaVAOneVisionEV,Reid2024Gemini1U,Li2022BLIPBL,Li2023BLIP2BL} offer such a pathway. These models integrate powerful language understanding with visual processing capabilities, enabling them to jointly reason about image content, textual descriptions, and quality attributes in a manner that surpasses simple pixel-level or feature distance comparisons. Their demonstrated proficiency in complex vision-language tasks, such as Visual Question Answering (VQA) and detailed image captioning~\cite{Li2022BLIPBL,Li2024LLaVAOneVisionEV,Li2023BLIP2BL}, highlights a capacity for sophisticated, nuanced understanding comparable to human judgment. It is precisely this ability to perform complex reasoning bridging vision and language that makes MLLMs uniquely qualified to evaluate immunization success according to our proposed criteria.

\begin{figure*}[t!]
  \centering
  % Replace with your actual figure file and path
  \includegraphics[width=0.95\textwidth]{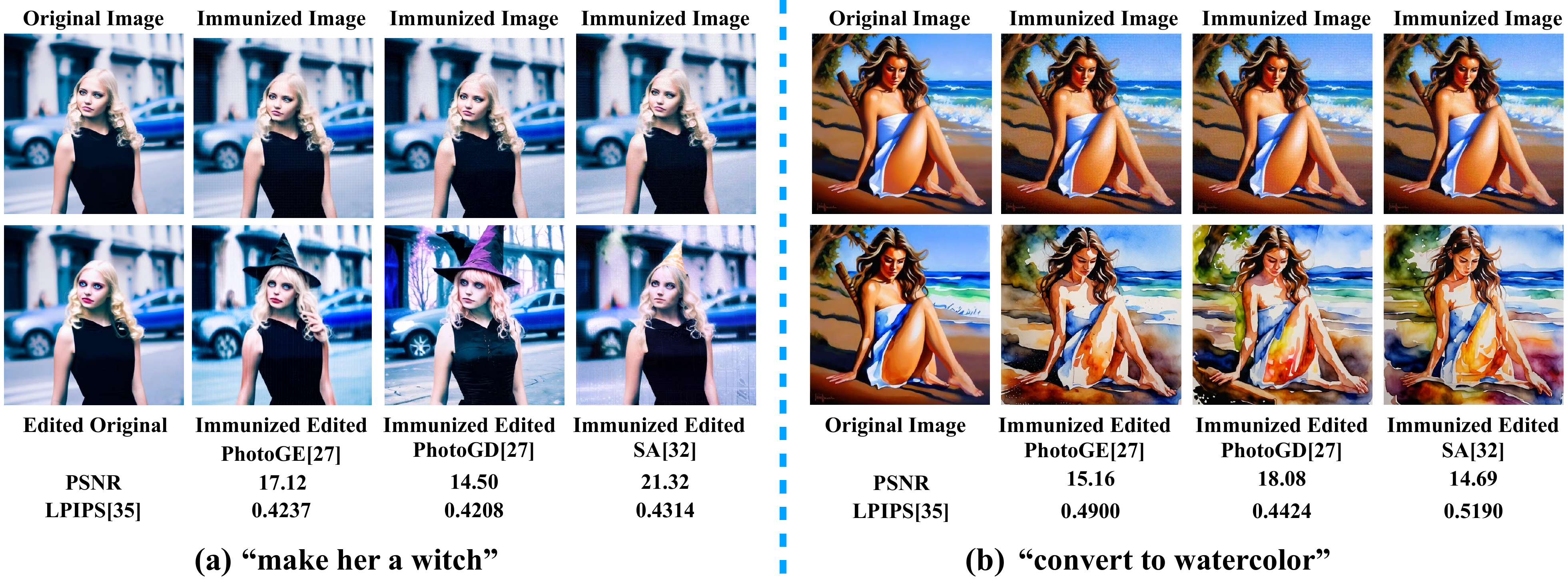} % Placeholder image
  \vspace{-5pt}
  \caption{Visual dissimilarity metrics fail to capture immunization success. For prompts like (a) ``make her a witch'' and (b) ``convert to watercolor'', edited immunized outputs exhibit significant visual deviations from original edits (high LPIPS and low PSNR) yet retain semantic alignment with the edit prompts. This demonstrates how traditional metrics conflate visual dissimilarity with functional immunization success, as the model still fulfills malicious editing. Notably, ``Immunized Edited'' refers to the edited output of the immunized image.}
  \label{fig:comp2}
\end{figure*}
\section{Preliminaries}
\subsection{Image Immunization Formulation.}
Let $f_\theta(\mathbf{x}_0, c)$ denote the output of a text-guided image editing model parameterized by $\theta$, given source image $\mathbf{x}_0$ and condition $c$. Image immunization aims to find an imperceptible perturbation $\boldsymbol{\delta}$, constrained within an $L_p$-norm ball $\|\boldsymbol{\delta}\|_p \leq \epsilon$ (usually $p=\infty$), such that the edited output $ f_\theta(\mathbf{x}_0 + \boldsymbol{\delta}, c)$ of the immunized image $\mathbf{x}_{\text{adv}} = \mathbf{x}_0 + \boldsymbol{\delta}$ differs significantly from the original edit $f_\theta(\mathbf{x}_0, c)$. Traditionally formulated through an adversarial lens~\cite{Goodfellow2014ExplainingAH, Madry2017TowardsDL}, the goal is to find $\boldsymbol{\delta}^*$ by solving:
\begin{equation}
\boldsymbol{\delta}^* = \underset{\|\boldsymbol{\delta}\|_p \leq \epsilon}{\arg\max}\, \mathcal{L}_{\text{dist}}\left( f_\theta(\mathbf{x}_0 + \boldsymbol{\delta}, c), f_\theta(\mathbf{x}_0, c) \right),
\label{eq:immunization_objective}
\end{equation}
where $\mathcal{L}_{\text{dist}}$ is a differentiable loss measuring the distance (e.g., $\ell_2$ distance) between the two edited outputs. This optimization is often performed using iterative gradient-based methods like Projected Gradient Descent (PGD)~\cite{Madry2017TowardsDL}. The PGD update at iteration $k$ is:
\begin{equation}
    \boldsymbol{\delta}_{k+1} = \Pi_{\|\cdot\|_p \leq \epsilon} \left( \boldsymbol{\delta}_{k} + \alpha \cdot g_k \right),
    \label{eq:pgd_update_immunization}
\end{equation}
where $\alpha$ is the step size, $g_k$ is the gradient direction (e.g., $\text{sign}(\nabla_{\boldsymbol{\delta}_{k}} \mathcal{L}_{\text{dist}})$), and $\Pi_{\|\cdot\|_p \leq \epsilon}(\cdot)$ projects the perturbation onto the $L_p$-ball. This traditional formulation, by maximizing distance, fundamentally overlooks the core requirement of image immunization, which is to disrupt semantic alignment with attacker intent, irrespective of deviation from any specific output.

\section{Rethinking Success Criteria for  Image Immunization}
\label{sec:isr_definition}

\subsection{Limitations of Existing Success Criteria}
\label{subsec:limitations_current_eval}

Current image immunization methods~\cite{Salman2023RaisingTC,Lo_2024_CVPR} usually evaluate the immunization performance by assessing the similarity between an edited immunized image and the corresponding edited original image, with lower computed similarity typically interpreted as indicative of superior immunization efficacy. PSNR and LPIPS~\cite{Zhang2018TheUE} are the typical metrics for pairwise image similarity or dissimilarity. For these metrics, a lower PSNR value and a higher LPIPS value respectively indicate better immunization.

However, this reliance on pairwise image similarity introduces fundamental limitations, as it does not directly address the specific objective of image immunization. Successful immunization requires preventing the editing model from executing the semantic intent of the edit prompt, not simply inducing visual deviations between an edited immunized image and the corresponding edited original image. Immunization may therefore fail even when substantial visual discrepancies exist, as long as the model successfully executes the edit’s semantic intent. This challenge is exacerbated by the inherent stochasticity of diffusion models and their ability to generate diverse outputs from identical prompts.

\begin{figure}{} 
  \centering 
  \includegraphics[width=0.33\textwidth]{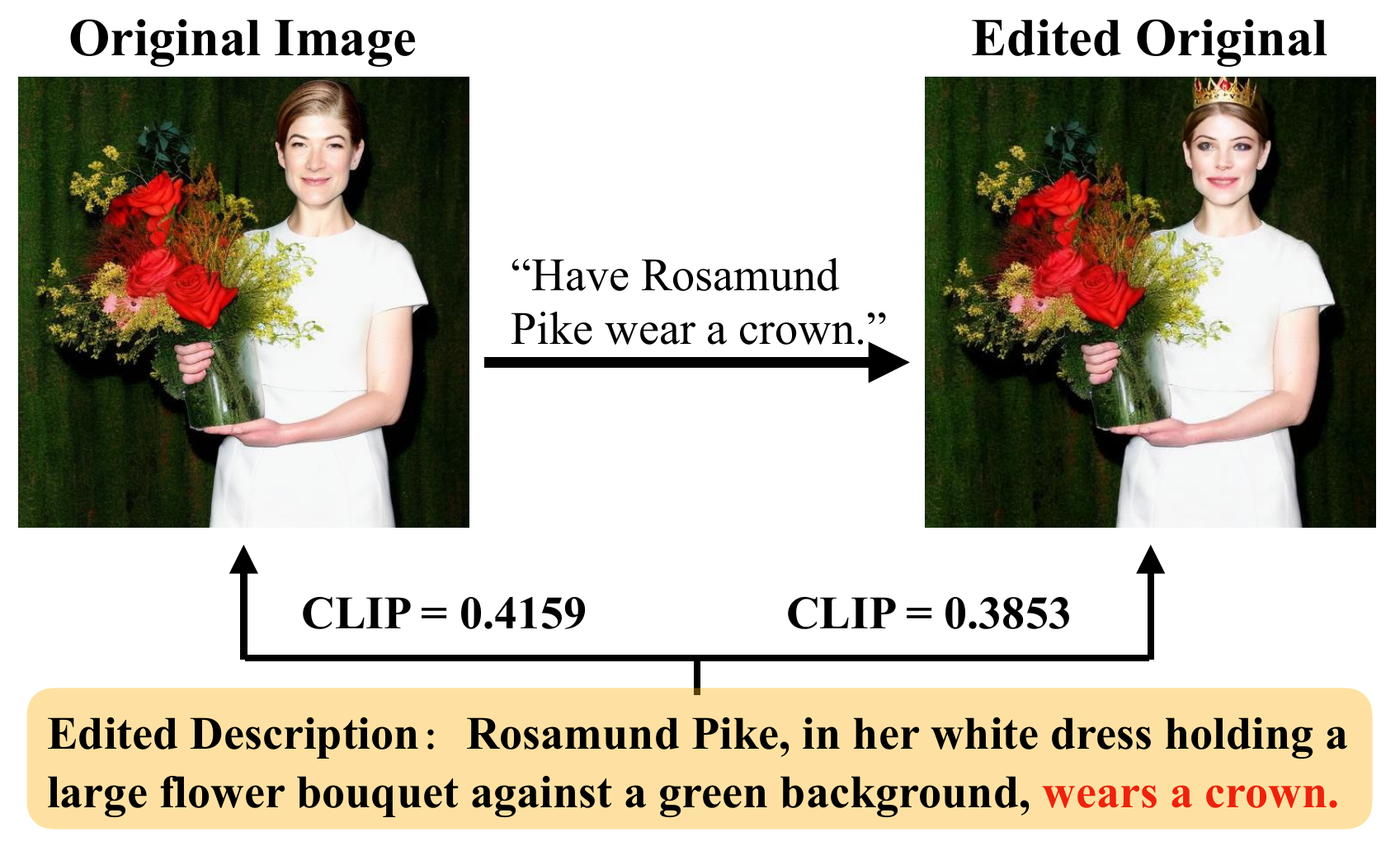} 
  \caption{\footnotesize Counterintuitive CLIP~\cite{Ramesh2022HierarchicalTI} scores reveal limitations for robust immunization assessment.}
  \label{fig:clip}
  
\end{figure}

Fig.~\ref{fig:comp2} intuitively illustrates the critical limitations of traditional metrics in evaluating image immunization. Consider the edit prompt ``make her a witch'' in Fig.~\ref{fig:comp2}(a), where the original edit yields a subject with heavy witch makeup. When images immunized by PhotoGD~\cite{Salman2023RaisingTC} and SA~\cite{Lo_2024_CVPR} undergo this edit, they exhibit drastically divergent PSNR scores, specifically 14.50 and 21.32, relative to the original edit. Despite this substantial numerical variance, both methods fail to achieve immunization. While their outputs differ visually from the reference by rendering features such as a pointed hat instead of makeup, they still faithfully execute the core semantic instruction to make the subject a witch. This disparity demonstrates that highly variable PSNR values and uniformly high LPIPS~\cite{Zhang2018TheUE} scores can accompany the same functional failure, which underscores the inadequacy of these metrics. They fail to reliably indicate true immunization success because the attacker's primary editing intent is achieved despite the metric readings. Similar observations regarding visual differences not equating to successful immunization apply to the ``convert to watercolor'' example shown in Fig.~\ref{fig:comp2}(b).

Furthermore, while models like CLIP~\cite{Ramesh2022HierarchicalTI} can assess the overall alignment between images and text, their evaluative capacity diminishes in scenarios such as image immunization, which necessitate precise judgment of fine-grained semantic changes. Fig.~\ref{fig:clip} clearly illustrates this limitation: an unedited original image, when matched against a detailed textual description of a target edited state (e.g., wears a crown), can paradoxically achieve a higher CLIP score than an image that has correctly executed this edit (i.e., actually depicts the crown). Such anomalous scoring suggests that CLIP may exhibit insufficient sensitivity in verifying the precise realization of specific, nuanced semantic elements, perhaps prioritizing the measurement of global visual-textual conceptual relatedness. Consequently, when evaluating whether image immunization has successfully thwarted a specific semantic manipulation, relying solely on CLIP scores may not provide a sufficiently reliable assessment.

Therefore, neither visual dissimilarity between the edited results of the immunized and original images, nor global image-text alignment scores, alone offers a reliable indication of successful immunization, thereby exposing the inadequacy of current evaluation standards predicated solely upon such metrics.

\subsection{Redefined Criteria for Successful Immunization}
\label{subsec:defining_true_success}

Given the inadequacy of pairwise similarity metrics highlighted in Section~\ref{subsec:limitations_current_eval}, we argue that defining immunization success necessitates assessing the utility of the final edited output $f_\theta(\mathbf{x_{adv}}, c)$ from the perspective of an adversary. An adversary typically seeks an output that adheres semantically to their prompt $c$ and preserves high visual fidelity. Therefore, immunization succeeds conceptually if it thwarts at least one of these adversarial goals.

We propose a revised definition where immunization is considered successful if applying the editing process $f_\theta$ to the immunized image $\mathbf{x_{adv}}$ with prompt $c$ yields an output $f_\theta(\mathbf{x_{adv}}, c)$ that satisfies at least one of the following conceptual conditions:

\begin{enumerate}
    \item \textbf{Semantic Mismatch.} The edited output $f_\theta(\mathbf{x_{adv}}, c)$ fails to align with the semantic intent conveyed by the text prompt $c$, thereby preventing the adversary from achieving their intended content modification.

    \item \textbf{Perceptual Degradation.} The edited output $f_\theta(\mathbf{x_{adv}}, c)$ exhibits substantial perceptual degradations including artifacts, distortions or noise that render it unacceptable or clearly manipulated. These degradations must be beyond any visual changes required by the edit prompt $c$ itself, making the output unusable for the adversary's purpose due to poor image quality.
\end{enumerate}
Crucially, immunization is deemed successful if it induces either semantic mismatches or perceptual degradations. Both outcomes prevent a malicious actor from obtaining their desired clean, semantically correct image reflecting prompt $c$. This redefinition of successful immunization directly addresses the limitations of prior evaluation paradigms  by focusing on the semantic correctness or image quality of $f_\theta(\mathbf{x}_{\text{adv}}, c)$ that determine its utility to a malicious actor.

\begin{algorithm}[t]
\caption{Synergistic Intermediate Feature Manipulation (SIFM)}
\label{alg:sifm_generation_condensed}
\KwIn{Input image $\mathbf{x}_{\text{orig}}$, prompt $\mathbf{c}$, diffusion model $f_\theta$, balance hyperparameter $\lambda$, budget $\epsilon$, step size $\alpha$, number of iterations $N$, set of diffusion timesteps $\mathcal{T}$}
\KwOut{Immunized image $\mathbf{x}_{\text{imu}}$}
\BlankLine
% Initialization
$\boldsymbol{\delta} \leftarrow \mathbf{0}$\;
$\mathbf{x}_{\text{imu}} \leftarrow \mathbf{x}_{\text{orig}}$\;
$\Phi_{\text{orig}} \leftarrow \{\phi_t(\mathbf{x}_{\text{orig}}, \mathbf{c}) \mid t \in \mathcal{T}\}$\; \quad // Pre-compute original features
\BlankLine
% PGD Iterations
\For{$n = 1$ \KwTo $N$}{
    $\mathbf{g}_{\text{total}} \leftarrow \mathbf{0}$\;
    % Accumulate gradients over timesteps
    \For{$t$ \textbf{in} $\mathcal{T}$}{
        $\phi_t^{\text{imu}} \leftarrow \phi_t(\mathbf{x}_{\text{imu}}, \mathbf{c})$\;
        $\mathcal{L}_t \leftarrow ||\phi_t^{\text{imu}}||_1^1 - \lambda \cdot \text{Dist}(\phi_t^{\text{imu}}, \Phi_{\text{orig}}[t])$\; \quad // SIFM loss
        $\mathbf{g}_{\text{total}} \leftarrow \mathbf{g}_{\text{total}} + \nabla_{\mathbf{x}_{\text{imu}}} \mathcal{L}_t$\;
    }
    % Average the accumulated gradients
    $\mathbf{g} \leftarrow \mathbf{g}_{\text{total}} / |\mathcal{T}|$\;
    
    % Update the perturbation delta via PGD
    $\boldsymbol{\delta} \leftarrow \text{clip}(\boldsymbol{\delta} - \alpha \cdot \text{sign}(\mathbf{g}), -\epsilon, \epsilon)$\;
    
    % Update the immunized image for the next iteration
    $\mathbf{x}_{\text{imu}} \leftarrow \text{clip}_{0,1}(\mathbf{x}_{\text{orig}} + \boldsymbol{\delta})$\;
}
\Return $\mathbf{x}_{\text{imu}}$\;
\end{algorithm}
\section{Synergistic Intermediate Feature Manipulation (SIFM)} % Or just Methodology
\label{sec:methodology}
Considering that the core criteria for successful immunization is preventing the editing model from strictly adhering to the text prompts, rather than merely inducing visual dissimilarity between an edited immunized image and edited original image, we propose \textbf{Synergistic Intermediate Feature Manipulation (SIFM)}. SIFM achieves effective immunization by manipulating intermediate features $\phi_t$ of the diffusion model at a set of specific timesteps $T = \{t_1, \dots, t_k\}$ using two synergistic objectives, detailed below. % Added "a set of" and "T = ..."

\textbf{Intermediate Feature Aggregation.} % New paragraph to define phi_t
The term $\phi_t(\mathbf{x}, c)$ represents an aggregation of features extracted from specific intermediate layers of the diffusion model's noise predictor network (e.g., U-Net or Diffusion Transformer) when processing image $\mathbf{x}$ with prompt $c$ at diffusion timestep $t$. Let $L_j(\mathbf{z}_t, c, t)$ denote the output of the $j$-th targeted intermediate layer within the noise predictor, where $\mathbf{z}_t$ is the noisy latent at timestep $t$. The aggregated feature representation $\phi_t(\mathbf{x}, c)$ is obtained by averaging the features from a predefined set of $M$ such targeted layers:
\begin{equation}
    \phi_t(\mathbf{x}, c) = \frac{1}{M} \sum_{j=1}^{M} L_j(\mathcal{E}(\mathbf{x}, t)_{\text{noisy}}, c, t)
    \label{eq:feature_aggregation}
\end{equation}
where $\mathcal{E}(\mathbf{x}, t)_{\text{noisy}}$ represents the process of obtaining the noisy latent representation of image $\mathbf{x}$ corresponding to timestep $t$. % Point to appendix for specific layer choices if needed

\subsection{Feature Distance Maximization for Semantic Mismatch}
\label{subsec:method_dist}
Prior research indicates that feature representations in a deep model encapsulate rich semantic information~\cite{Naseer2021IntriguingPO,Raghu2021DoVT}. In image editing models, deeper layers of the noise predictor exhibit stronger alignment between textual instructions and intermediate image features, progressively refining outputs to reflect the semantic intent of the edit prompt. Building on this observation, we maximize the semantic divergence between the immunized image's intermediate features $\phi_t(\mathbf{x}_0 + \delta, c)$ and those of the original image $\phi_t(\mathbf{x}_0, c)$ at each sampled timestep $t \in T$. By perturbing these semantically critical features, we disrupt the model’s ability to synthesize outputs that adhere to the edit prompt, thereby approaching effective image immunization.

Specifically, for each $t \in T$, we define a distance-maximizing loss component: % Removed "Specifically, let phi_t(x,c) denote..." as it's now defined above
\begin{equation}
    \mathcal{L}_{\text{dist}}(\delta; t) = \text{Dist}(\phi_t(\mathbf{x}_0 + \delta, c), \phi_t(\mathbf{x}_0, c))
    \label{eq:loss_dist_t}
\end{equation}
Herein, $\text{Dist}(\cdot, \cdot)$ represents a distance metric (e.g., MSE) applied to feature representations. By maximizing this objective at each selected timestep during $\delta$ optimization, we shift the diffusion trajectory of the immunized image away from that of the original image under identical edit conditioning. This promotes semantic divergence, increasing the likelihood that the final immunized edit will semantically mismatch the edit prompt.

\subsection{Feature Norm Minimization for perceptual degradation}
\label{subsec:method_norm}
The feature layer representations $\phi_t(\mathbf{x}_0 + \delta, c)$ defined in Eq.~\eqref{eq:feature_aggregation} serve as the semantic and structural blueprint for image synthesis. To induce perceptual degradations, our strategy is not merely to reduce the magnitude of these features, but to systematically impoverish them by enforcing sparsity. We achieve this by minimizing their L1 norm, which acts as a powerful sparsity-inducing regularizer. This objective compels the model to discard less critical feature activations, effectively creating a compressed and lossy representation. When the diffusion model attempts to generate an image from this sparse blueprint, the amplified loss of information undermines its ability to reconstruct details with precision or preserve visual coherence.

Specifically, our objective is formulated as an L1 norm-minimizing loss component applied at each chosen timestep $t \in T$:
\begin{equation}
    \mathcal{L}_{\text{norm}}(\delta; t) = ||\phi_t(\mathbf{x}_0 + \delta, c)||_1
    \label{eq:loss_norm_t}
\end{equation}
By attenuating and zeroing out feature activations, this L1 penalty systematically disrupts the feature refinement process during the reverse sampling steps. The initial loss of detail propagates and magnifies through subsequent layers, amplifying reconstruction errors and noise accumulation. This process reliably cascades into the structural incoherence and disruptive artifacts in the final output that fulfill our perceptual degradation criterion for successful immunization, rendering the edited images unusable.

\subsection{Combined Objective and Perturbation Generation} % Changed title in previous iteration
\label{subsec:method_combined}

To effectively leverage both pathways to immunization success, SIFM combines the two objectives synergistically across the set of chosen timesteps $T$. For each timestep $t \in T$, a per-timestep loss $\mathcal{L}_{\text{SIFM}}(\delta; t)$ is formulated. We optimize the perturbation $\delta$ by minimizing an aggregation of these per-timestep losses. The per-timestep loss involves minimizing $\mathcal{L}_{\text{norm}}(\delta; t)$ while simultaneously maximizing $\mathcal{L}_{\text{dist}}(\delta; t)$:
\begin{equation}
    \mathcal{L}_{\text{SIFM}}(\delta; t) = \mathcal{L}_{\text{norm}}(\delta; t) - \lambda \cdot \mathcal{L}_{\text{dist}}(\delta; t)
    \label{eq:loss_sifm_t}
\end{equation}
Here, $\lambda > 0$ is a hyperparameter. The overall SIFM objective $\mathcal{L}_{\text{SIFM}}^{\text{total}}(\delta)$ is then the average of these per-timestep losses:
\begin{equation}
    \mathcal{L}_{\text{SIFM}}^{\text{total}}(\delta) = \frac{1}{|T|} \sum_{t \in T} \mathcal{L}_{\text{SIFM}}(\delta; t)
    \label{eq:loss_sifm_total}
\end{equation}
We generate the perturbation $\delta$ by iteratively minimizing Eq.~\eqref{eq:loss_sifm_total} using Projected Gradient Descent (PGD), as detailed in Algorithm~\ref{alg:sifm_generation_condensed}.
\begin{figure*}[t!]
  \centering
  \includegraphics[width=0.9\textwidth]{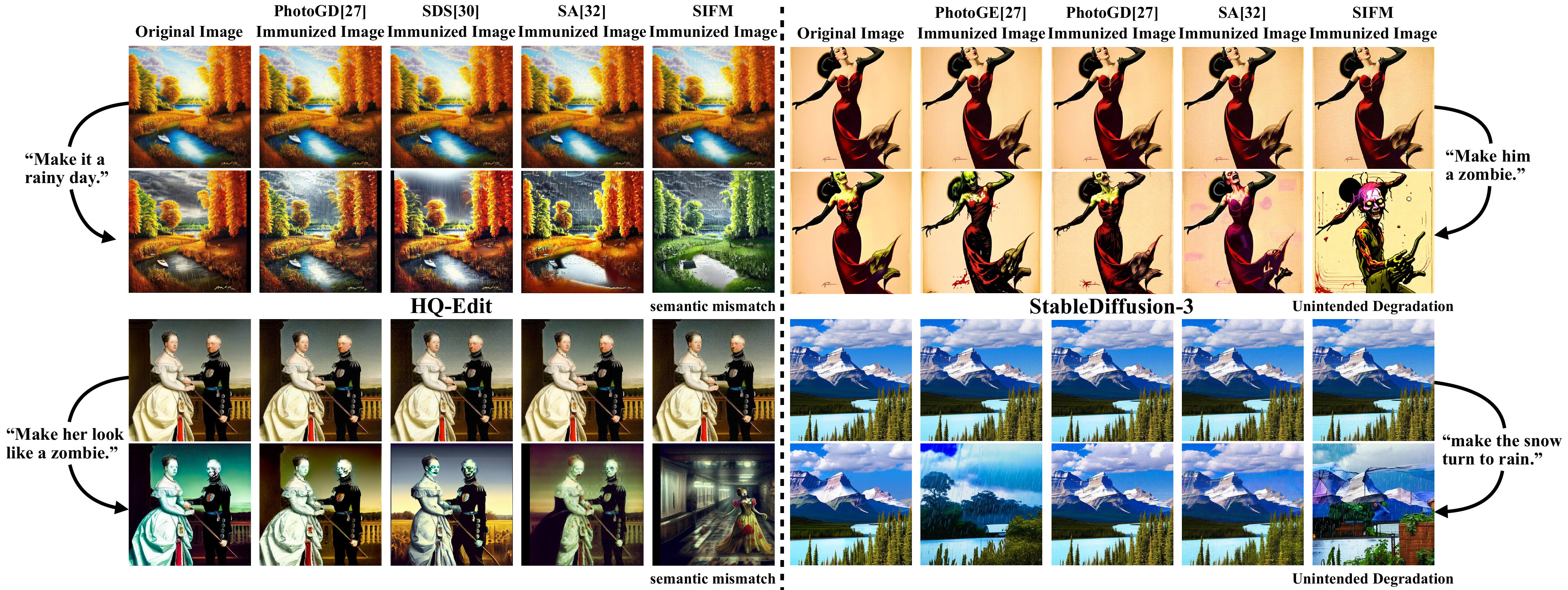} % 替换为您的图片文件名
  \vspace{-5pt}
  \caption{Qualitative results on HQ-Edit~\cite{Hui2024HQEditAH} and StableDiffusion-3~\cite{Esser2024ScalingRF}. SIFM (rightmost) successfully immunizes by making semantic mismatch to edit prompts or perceptual degradation, whereas baselines not.}
  \label{fig:r1}
\end{figure*}

\section{Immunization Success Rate}
\label{subsec:defining_and_quantifying_true_success}
To rigorously evaluate immunization efficacy under our redefined criteria, we propose the \textbf{Immunization Success Rate (ISR)}. Unlike traditional metrics (PSNR, LPIPS~\cite{Zhang2018TheUE}), which measure superficial pixel- or feature-level deviations, the ISR directly quantifies whether edited outputs violate at least one objective: semantic mismatches or perceptual degradations. This aligns with our immunization success criteria, addressing the limitations of prior metrics that conflate visual dissimilarity with image editing failure.

The ISR leverages Multimodal Large Language Models (MLLMs), i.e., Gemini 2.5 Pro~\cite{Comanici2025Gemini2P} and Gemini 2.5 Flash~\cite{Comanici2025Gemini2P}, to automatically evaluate whether an immunized edit $f_\theta(\mathbf{x}_{\text{adv}}, c)$ violates either semantic alignment (deviation from the edit prompt) or perceptual degradations (degradations unrelated to the edit prompt). To ensure a conservative and high-confidence measure, we aggregate their verdicts using a strict agreement policy. An immunization attempt is logged as a success only if both MLLMs independently conclude that it has successfully failed the edit by inducing a semantic failure or significant degradation. If at least one model judges the attempt as a failure, the sample is conservatively counted as an immunization failure. This methodology provides a robust lower bound on the perceived immunization effectiveness by minimizing potential false positives from any single model.

We define $N_{\text{success}}$ as the number of samples where this MLLM consensus confirms immunization success, and $N_{\text{total}}$ as the total number of evaluated samples. The ISR is then computed as:
\begin{equation}
\label{eq:isr_calculation} % You can change this label if needed
ISR = \frac{N_{\text{success}}}{N_{\text{total}}}
\end{equation}

\section{Experiments}

Comprehensive qualitative and quantitative evaluations were conducted in this Section. All methods were constrained to a perturbation budget of $\epsilon=0.03$ and limited to 100 optimization iterations.

\subsection{Models and Datasets}
\textbf{Models and Datasets}
We evaluated immunization methods on StableDiffusion-3~\cite{Esser2024ScalingRF}, HQ-Edit~\cite{Hui2024HQEditAH}, and Instructpix2pix~\cite{Brooks2022InstructPix2PixLT}. Due to the absence of standardized benchmarks, we created a custom dataset of 100 images (35 portraits, 35 landscapes, 30 artworks) selected from Instructpix2pix-clip-filtered~\cite{instructpix2pix-clip-filtered}. Each image's original prompt from this collection was used for perturbation generation.
Acknowledging the unpredictability of attacker prompts, we generated five novel, semantically distinct editing prompts per image for cross-prompt generalization assessment. This yielded 500 unique scenarios per editing model to evaluate generalization.

\textbf{Evaluation Metrics.}
Our evaluation utilizes two metric categories. The first category is the traditional metrics, including PSNR, SSIM~\cite{1284395}, VIFp~\cite{1576816}, FSIM~\cite{5705575} and LPIPS~\cite{Zhang2018TheUE}, quantify perceptual differences between edited immunized image and edited original images. The second one is our proposed \textbf{Immunization Success Rate (ISR)}. As established in Section~\ref{subsec:defining_and_quantifying_true_success}, ISR measures immunization success via either semantic mismatches with the prompt or significant perceptual degradations, assessed by MLLMs (Gemini 2.5 Pro~\cite{Comanici2025Gemini2P}, Gemini 2.5 Flash~\cite{Comanici2025Gemini2P}) and calculated by Eq~\eqref{eq:isr_calculation}. A higher ISR signifies more effective immunization. 

\begin{table}[t!]
\centering
\caption{Consistency analysis of MLLM judgments (Gemini 2.5 Flash vs. Pro) against human evaluators. The high human to MLLM consistency validates ISR as a reliable proxy for human perception.}
\label{tab:human_study}
\resizebox{\columnwidth}{!}{%
\begin{tabular}{@{}lccc@{}}
\toprule
\textbf{Consistency Metric} & \textbf{Semantic Mismatch} & \textbf{Quality Degradation} & \textbf{Final Success (ISR)} \\
\midrule
Human to Human  & 84\% & 79\% & 76\% \\
MLLM to MLLM & 80\% & 70\% & 74\% \\
Human to MLLM  & 73\% & 72\% & 74\% \\
\bottomrule
\end{tabular}%
}
\end{table}

\textbf{Validation of the ISR Metric via Human Study.}
To rigorously validate the reliability of the MLLM based ISR metric, we conducted a formal human evaluation study. We recruited three independent human evaluators to assess a representative subset of 150 samples selected via stratified random sampling across all methods, prompt types, and image categories. For each sample (Original Image, Edited Immunized Image, Edit Prompt), evaluators provided binary judgments on semantic mismatch, quality degradation, and overall immunization success, following the same logic as the MLLMs. We then measured the consistency between human evaluators (Human to Human), between our two MLLMs: Gemini 2.5 Pro and Gemini 2.5 Flash (MLLM to MLLM), and, most importantly, between human and MLLM judgments (Human to MLLM).

The results are summarized in Tab.~\ref{tab:human_study}. The average Human to MLLM consistency for the final success judgment is 74\%, which is remarkably close to the inter human agreement rate of 76\%. This high degree of alignment provides strong empirical evidence that our Gemini based ISR metric serves as a reliable and accurate proxy for human perception of immunization success, justifying its use as our primary evaluation metric.

\begin{table*}[t!]
\vspace{-12pt}
\centering
\caption{Quantitative performance of immunization methods on StableDiffusion-3~\cite{Esser2024ScalingRF} using original and unseen prompts. Immunized images were evaluated by editing them using the original prompt employed during optimization and five unseen prompts for generalization assessment.}
\label{tab:sd3_original_unseen}
\resizebox{\textwidth}{!}{%
\begin{tabular}{l cccccc cccccc }
\toprule
Method & \multicolumn{6}{c}{Original Prompt} & \multicolumn{6}{c}{Unseen Prompts} \\
\cmidrule(lr){2-7} \cmidrule(lr){8-13}
& PSNR $\downarrow$ & SSIM~\cite{1284395} $\downarrow$ & VIFp~\cite{1576816} $\downarrow$ & FSIM~\cite{5705575} $\downarrow$ & LPIPS~\cite{Zhang2018TheUE} $\uparrow$ & ISR $\uparrow$ & PSNR $\downarrow$ & SSIM~\cite{1284395} $\downarrow$ & VIFp~\cite{1576816} $\downarrow$ & FSIM~\cite{5705575} $\downarrow$ & LPIPS~\cite{Zhang2018TheUE} $\uparrow$ & ISR $\uparrow$ \\
\midrule
PhotoGE~\cite{Salman2023RaisingTC} & 18.14 & 0.5940 & 0.2076 & 0.7921 & 0.3796 & 63\% & 15.94 & 0.5084 & 0.1391 & 0.7348 & 0.4587 & 54\% \\
PhotoGD~\cite{Salman2023RaisingTC} & 18.43 & 0.5809 & 0.2150 & 0.7890 & 0.3921 & 67\% & 17.59 & 0.5439 & 0.1568 & 0.7641 & 0.4373 & 48\% \\
ED~\cite{Chen2024EditShieldPU}      & 22.88 & 0.7600 & 0.3894 & 0.8834 & 0.2155 & 50\% & 20.03 & 0.6832 & 0.2889 & 0.8345 & 0.2817 & 26\% \\
SA~\cite{Lo_2024_CVPR}      & 17.85 & 0.5583 & 0.1894 & 0.7643 & 0.4225 & 70\% & 16.73 & 0.5213 & 0.1462 & 0.7403 & 0.4862 & 48\% \\
\textbf{SIFM} & \textbf{15.79} & \textbf{0.4747} & \textbf{0.1237} & \textbf{0.7162} & \textbf{0.5046} & \textbf{79\%} & \textbf{15.50} & \textbf{0.4703} & \textbf{0.1380} & \textbf{0.7174} & \textbf{0.5042} & \textbf{65\%} \\
\bottomrule
\end{tabular}%
}
\end{table*}
\begin{table*}[t!]
\vspace{-12pt}
\centering
\caption{Quantitative performance of immunization methods on Instructpix2pix~\cite{Brooks2022InstructPix2PixLT} using original and unseen prompts. Immunized images were evaluated by editing them using the original prompt employed during optimization and five unseen prompts for generalization assessment.}
\label{tab:ip2p_original_unseen}
\resizebox{\textwidth}{!}{%
\begin{tabular}{l cccccc cccccc }
\toprule
Method & \multicolumn{6}{c}{Original Prompt} & \multicolumn{6}{c}{Unseen Prompts} \\
\cmidrule(lr){2-7} \cmidrule(lr){8-13}
& PSNR $\downarrow$ & SSIM~\cite{1284395} $\downarrow$ & VIFp~\cite{1576816} $\downarrow$ & FSIM~\cite{5705575} $\downarrow$ & LPIPS~\cite{Zhang2018TheUE} $\uparrow$ & ISR $\uparrow$ & PSNR $\downarrow$ & SSIM~\cite{1284395} $\downarrow$ & VIFp~\cite{1576816} $\downarrow$ & FSIM~\cite{5705575} $\downarrow$ & LPIPS~\cite{Zhang2018TheUE} $\uparrow$ & ISR $\uparrow$ \\
\midrule
PhotoGE~\cite{Salman2023RaisingTC} & 16.15 & 0.5945 & 0.1858 & 0.7849 & 0.3801 & 75\% & 16.02 & 0.5699 & 0.1943 & 0.7773 & 0.4260 & 51\% \\
PhotoGD~\cite{Salman2023RaisingTC} & 17.83 & 0.6658 & 0.2408 & 0.8233 & 0.3157 & 69\% & 16.91 & 0.6122 & 0.2281 & 0.7963 & 0.3769 & 55\% \\
ED~\cite{Chen2024EditShieldPU}      & 19.04 & 0.7239 & 0.3108 & 0.8469 & 0.3070 & 48\% & 16.99 & 0.6131 & 0.2370 & 0.7968 & 0.3703 & 36\% \\
ACE~\cite{Zheng2023TargetedAI}     & 17.01 & 0.6219 & 0.1989 & 0.7963 & 0.4258 & 68\% & 16.42 & 0.5784 & 0.1927 & 0.7795 & 0.4159 & 44\% \\
SDS~\cite{Xue2023TowardEP}     & 15.40 & 0.5954 & 0.1836 & 0.7669 & 0.4457 & 69\% & 15.80 & 0.5469 & 0.1703 & 0.7585 & 0.4444 & 52\% \\
MIST~\cite{Liang2023MistTI}    & 16.34 & 0.6265 & 0.2283 & 0.7943 & 0.4420 & 64\% & 16.31 & 0.5755 & 0.1985 & 0.7791 & 0.4192 & 42\% \\
SA~\cite{Lo_2024_CVPR}      & 16.53 & 0.5961 & 0.1939 & 0.8014 & 0.3986 & 66\% & 16.15 & 0.5719 & 0.1923 & 0.7775 & 0.4210 & 46\% \\
\textbf{SIFM} & \textbf{14.94} & \textbf{0.5330} & \textbf{0.1584} & \textbf{0.7253} & \textbf{0.4497} & \textbf{83\%} & \textbf{15.11} & \textbf{0.5296} & \textbf{0.1691} & \textbf{0.7433} & \textbf{0.4595} & \textbf{68\%} \\
\bottomrule
\end{tabular}%
}
\end{table*}

\begin{table*}[t!]
\centering
\caption{Quantitative performance of immunization methods on HQ-Edit~\cite{Hui2024HQEditAH} using original and unseen prompts. Immunized images were evaluated by editing them using the original prompt employed during optimization and five unseen prompts for generalization assessment.}
\label{tab:hqedit_original_unseen}
\resizebox{\textwidth}{!}{%
\begin{tabular}{l cccccc cccccc }
\toprule
Method & \multicolumn{6}{c}{Original Prompt} & \multicolumn{6}{c}{Unseen Prompts} \\
\cmidrule(lr){2-7} \cmidrule(lr){8-13}
& PSNR $\downarrow$ & SSIM~\cite{1284395} $\downarrow$ & VIFp~\cite{1576816} $\downarrow$ & FSIM~\cite{5705575} $\downarrow$ & LPIPS~\cite{Zhang2018TheUE} $\uparrow$ & ISR $\uparrow$ & PSNR $\downarrow$ & SSIM~\cite{1284395} $\downarrow$ & VIFp~\cite{1576816} $\downarrow$ & FSIM~\cite{5705575} $\downarrow$ & LPIPS~\cite{Zhang2018TheUE} $\uparrow$ & ISR $\uparrow$ \\
\midrule
PhotoGE~\cite{Salman2023RaisingTC} & 10.23 & 0.2968 & 0.0528 & 0.5924 & 0.6261 & 84\% & 8.89 & 0.2772 & 0.0470 & 0.5748 & 0.6425 & 62\% \\
PhotoGD~\cite{Salman2023RaisingTC} & 10.87 & 0.2956 & 0.0504 & 0.5929 & 0.6384 & 88\% & 9.01 & 0.2664 & 0.0435 & 0.5752 & 0.6451 & 51\% \\
ED~\cite{Chen2024EditShieldPU}      & 9.90 & 0.3046 & 0.0605 & 0.6012 & 0.5929 & 88\% & 9.08 & 0.2822 & 0.0499 & 0.5781 & 0.6312 & 45\% \\
ACE~\cite{Zheng2023TargetedAI}     & 9.60 & 0.3046 & 0.0477 & 0.5983 & 0.6393 & 92\% & 8.99 & 0.2737 & 0.0453 & 0.5764 & 0.6430 & 58\% \\
SDS~\cite{Xue2023TowardEP}     & 9.37 & 0.2867 & 0.0545 & 0.5884 & 0.6366 & 91\% & 8.90 & 0.2613 & 0.0430 & 0.5722 & 0.6459 & 69\% \\
MIST~\cite{Liang2023MistTI}    & 9.89 & 0.3311 & 0.0564 & 0.6037 & 0.6394 & 88\% & 9.22 & 0.2835 & 0.0485 & 0.5809 & 0.6386 & 59\% \\
SA~\cite{Lo_2024_CVPR}      & 10.02 & 0.2886 & 0.0495 & 0.5905 & 0.6443 & 90\% & 8.90 & 0.2624 & 0.0426 & 0.5714 & 0.6487 & 52\% \\
\textbf{SIFM} & \textbf{9.23} & \textbf{0.2733} & \textbf{0.0457} & \textbf{0.5860} & \textbf{0.6719} & \textbf{97\%} & \textbf{8.47} & \textbf{0.2553} & \textbf{0.0374} & \textbf{0.5711} & \textbf{0.6599} & \textbf{71\%} \\
\bottomrule
\end{tabular}%
}
\end{table*}
\begin{table*}[t!]
\centering
\caption{Cross model transferability results. Perturbations are generated on a source model and tested against a different, unseen target model. Results are evaluated using the human validated ISR.}
\label{tab:transferability}
\resizebox{\textwidth}{!}{%
\begin{tabular}{l cccccc cccccc }
\toprule
Method & \multicolumn{6}{c}{Instructpix2pix $\rightarrow$ StableDiffusion-3} & \multicolumn{6}{c}{StableDiffusion-3 $\rightarrow$ Instructpix2pix} \\
\cmidrule(lr){2-7} \cmidrule(lr){8-13}
& PSNR $\downarrow$ & SSIM $\downarrow$ & LPIPS $\uparrow$ & VIFp $\downarrow$ & FSIM $\downarrow$ & ISR $\uparrow$ & PSNR $\downarrow$ & SSIM $\downarrow$ & LPIPS $\uparrow$ & VIFp $\downarrow$ & FSIM $\downarrow$ & ISR $\uparrow$ \\
\midrule
PhotoGE & 20.76 & 0.6578 & 0.3222 & 0.2467 & 0.8407 & 37\% & 18.50 & 0.6597 & 0.3378 & 0.2333 & 0.8311 & 51\% \\
PhotoGD & 22.08 & 0.7276 & 0.2453 & 0.3168 & 0.8690 & 31\% & 17.77 & 0.6324 & 0.3466 & 0.2261 & 0.8157 & 52\% \\
SA      & 20.98 & 0.6516 & 0.3492 & 0.2419 & 0.8438 & 42\% & 18.60 & 0.6622 & 0.3457 & 0.2289 & 0.8359 & 50\% \\
\textbf{SIFM} & \textbf{21.14} & \textbf{0.6569} & \textbf{0.3496} & \textbf{0.2411} & \textbf{0.8412} & \textbf{49\%} & \textbf{17.51} & \textbf{0.6426} & \textbf{0.3548} & \textbf{0.2295} & \textbf{0.8305} & \textbf{62\%} \\
\bottomrule
\end{tabular}%
}
\end{table*}

\begin{table*}[h]
\centering
\caption{Ablation study of SIFM's objective function components. The proposed SIFM incorporates both $\mathcal{L}_{\text{dist}}$ and $\mathcal{L}_{\text{norm}}$.}
\label{tab:component_ablation}
\begin{tabular}{@{}l cc cccccc@{}} 
\toprule

\multirow{2}{*}{\textbf{Method}} & \multicolumn{2}{c}{\textbf{Components}} & \multicolumn{6}{c}{\textbf{Metrics}} \\
\cmidrule(lr){2-3} \cmidrule(l){4-9}

 & $\mathcal{L}_{\text{dist}}$ & $\mathcal{L}_{\text{norm}}$ & PSNR$\downarrow$ & SSIM$\downarrow$ & LPIPS$\uparrow$ & VIFp$\downarrow$ & FSIM$\downarrow$ & \textbf{ISR}$\uparrow$ \\
\midrule

- & \checkmark & & 18.20 & 0.5312 & 0.4394 & 0.1965 & 0.8130 & 66\% \\

- & & \checkmark & 16.41 & 0.4827 & 0.4916 & 0.1367 & 0.7275 & 73\% \\

\textbf{SIFM} & \checkmark & \checkmark & \textbf{15.79} & \textbf{0.4747} & \textbf{0.5046} & \textbf{0.1237} & \textbf{0.7162} & \textbf{79\%} \\

\bottomrule
\end{tabular}
\end{table*}

\begin{table*}[h] 
  \centering
  \caption{Ablation study on the hyperparameter $\lambda$, which balances $\mathcal{L}_{\text{norm}}$ and $\mathcal{L}_{\text{dist}}$. The evaluation was performed on StableDiffusion-3.}
  \label{tab:lambda_ablation} % More descriptive label
  \begin{tabular}{@{}lcccccc@{}} 
    \toprule
    Value of $\lambda$ & PSNR$\downarrow$ & SSIM$\downarrow$ & LPIPS$\uparrow$ & VIFp$\downarrow$ & FSIM$\downarrow$ & \textbf{ISR}$\uparrow$ \\ 
    \midrule
    0.001   & 15.85 & 0.4832 & 0.4956 & 0.1287 & 0.7220 & 73\% \\
    0.01    & 15.77 & 0.4769 & 0.4954 & 0.1242 & 0.7213 & 76\% \\
    \textbf{0.1} & 15.79 & 0.4747 & \textbf{0.5046} & \textbf{0.1237} & \textbf{0.7162} & \textbf{79\%} \\
    1.0     & 15.78 & 0.4820 & 0.4915 & 0.1284 & 0.7193 & 71\% \\
    10.0    & 15.81 & 0.4752 & 0.4938 & 0.1246 & 0.7236 & 70\% \\
    100.0   & \textbf{15.76} & \textbf{0.4749} & 0.4974 & 0.1239 & 0.7234 & 71\% \\
    1000.0  & 15.83 & 0.4760 & 0.4935 & 0.1250 & 0.7236 & 72\% \\
    \bottomrule
  \end{tabular}
\end{table*}

\subsection{Experimental Results}
\subsubsection{White Box Immunization Evaluation}
We begin by assessing the performance in a white-box scenario, where immunization perturbations are optimized specifically for the target editing models, ensuring a rigorous evaluation of defense capabilities.

\textbf{Immunization Results on StableDiffusion-3.}
We first evaluated our method on the state of the art StableDiffusion-3~\cite{Esser2024ScalingRF} model. Due to its novel Diffusion Transformer (DiT) architecture, several baselines (ACE, MIST, SDS) were incompatible and thus excluded from this evaluation. As shown in Tab.~\ref{tab:sd3_original_unseen}, SIFM demonstrates superior performance under the original editing prompts. It achieves an ISR of 79\%, significantly outperforming the next best method, SA, by 9\%. This superior immunization success is also reflected in traditional metrics, where SIFM achieves the lowest PSNR of 15.79 and the highest LPIPS of 0.5046, indicating maximal disruption at a perceptual level. Qualitative results in Fig.~\ref{fig:r1} illustrates this. For the prompt ``Make him a zombie'', baselines successfully fulfill the malicious intent by creating a zombie like figure, whereas SIFM causes a complete unintended degradation, making the edited image unusable for the attackers. When tested for generalization on unseen prompts, SIFM maintains its clear lead with an ISR of 65\%.

\textbf{Immunization Results on HQ-Edit.}
On the HQ-Edit~\cite{Hui2024HQEditAH} model, SIFM establishes a dominant lead, achieving a near perfect ISR of 97\% on original prompts, surpassing the strongest baseline ACE by 5 percentage points which is shown in Tab.~\ref{tab:hqedit_original_unseen}. This exceptional success rate is complemented by its top performance across all traditional metrics, including the lowest PSNR 9.23 and the highest LPIPS 0.6719, demonstrating both comprehensive semantic and perceptual disruption. The qualitative results in Fig.~\ref{fig:r1} provide clear evidence of its effectiveness. For example, when prompted to ``Make her look like a zombie'', baselines correctly apply the zombie effect to the subjects in the portrait. In stark contrast, SIFM thwarts this malicious edit by forcing the model to generate an entirely different scene: a hallway, thereby completely derailing the edit's intent. For unseen prompts, it again proves its superior generalization with a top performing ISR of 71\%.

\textbf{Immunization Results on Instructpix2pix.}
Finally, on Instructpix2pix~\cite{Brooks2022InstructPix2PixLT}, SIFM continues its strong performance as shown in Tab.~\ref{tab:ip2p_original_unseen}. Under original prompts, it achieves the highest ISR of 83\%, a significant margin over the runner up, PhotoGE's 75\%. This is corroborated by its leading scores in traditional metrics, where it records the lowest PSNR of 14.94 and the highest LPIPS of 0.4497. This evaluation also highlights the necessity of the ISR metric over traditional ones. For instance, while MIST achieves a higher LPIPS of 0.4420 than ACE's 0.4258, suggesting greater perceptual difference, its actual immunization success as measured by ISR is lower (64\% for MIST vs. 68\% for ACE). This discrepancy demonstrates that high perceptual distance does not always translate to successful neutralization of malicious intent, reinforcing the importance of our semantically grounded ISR metric. The method also excels in generalization, securing a leading ISR of 68\% on unseen prompts and confirming its efficacy is broadly applicable.

\subsubsection{Black Box Immunization Evaluation}

A critical aspect of a practical immunization method is its ability to generalize to unseen target models, a scenario that relaxes the white box assumption. To this end, we conducted cross model transferability experiments where perturbations generated on a source model were used to immunize images against edits by a different, unseen target model. We evaluated two scenarios: (1) Instructpix2pix $\rightarrow$ StableDiffusion-3, and (2) StableDiffusion-3 $\rightarrow$ Instructpix2pix.

The results in Table~\ref{tab:transferability}, demonstrate SIFM's promising transferability. While performance on traditional metrics like PSNR and LPIPS is comparable across other methods, SIFM exhibits a decisive advantage in ISR, our primary metric for immunization success. In the INS $\rightarrow$ SD3 scenario, SIFM achieves the highest ISR of 49\%. More notably, in the SD3 $\rightarrow$ INS transfer scenario, SIFM achieves an ISR of \textbf{62\%}, outperforming the next best method by a significant margin of 10 percentage points. This suggests that by targeting fundamental intermediate features, SIFM's perturbations are more robust and less model specific, granting them superior generalization capabilities in a more realistic black box setting.

\subsection{Ablation Studies}
We conducted comprehensive ablation studies to validate the effectiveness of SIFM's core components and to determine the optimal setting for its key hyperparameter, $\lambda$.

\textbf{Effectiveness of SIFM Components.}
First, an objective function ablation highlights the synergistic effect of combining semantic mismatch induction via feature distance maximization ($\mathcal{L}_{\text{dist}}$) and perceptual degradations induction via feature norm minimization ($\mathcal{L}_{\text{norm}}$). As shown in Tab.~\ref{tab:component_ablation}, the full SIFM framework achieves a superior ISR of 79\%, significantly outperforming the individual objectives. This result empirically validates that our dual-objective approach is more effective than targeting either failure mode alone.

% Note: I've updated the table label to be more descriptive.
% Please replace your existing table with this one.

\textbf{Impact of Hyperparameter $\lambda$.}
Second, we analyzed the impact of the hyperparameter $\lambda$, which balances the $\mathcal{L}_{\text{norm}}$ and $\mathcal{L}_{\text{dist}}$ components. To determine its optimal setting, we evaluated SIFM's performance on StableDiffusion-3 across several orders of magnitude for $\lambda$.

The results are presented in Tab.~\ref{tab:lambda_ablation}. Setting $\lambda=0.1$ achieves the highest Immunization Success Rate (ISR) at \textbf{79\%}. While this value does not yield the absolute best score for each single traditional metric (e.g., $\lambda=100.0$ shows a slightly lower PSNR), its performance on these metrics is highly competitive, and it secures the peak LPIPS score. Considering the significant peak in our primary evaluation criterion, ISR, coupled with strong results across secondary metrics, we determined that $\lambda=0.1$ represents the best overall trade-off. Consequently, this value was used for all other experiments reported in this paper.

\section{Conclusion}
\label{sec:conclusion}

This work redefines image immunization success by shifting focus from traditional pairwise image dissimilarity metrics between edited immunized images and the original edited results to adversarial utility: an edit is neutralized if the output violates semantic alignment with the edit prompt or incurs perceptual degradations unrelated to the edit instruction. Our SIFM engineers these failure modes through synergistic manipulation of intermediate diffusion features: maximizing feature distance to induce semantic mismatches and minimizing feature norms to degrade perceptual integrity. Immunization efficacy is rigorously quantified via the ISR, an MLLM-driven metric that evaluates adherence to our criteria. Extensive experiments demonstrate SIFM’s superiority, achieving the state-of-the-art results for image immunization.

\bibliographystyle{IEEEtran}

\bibliography{main}

\vspace{-22pt}
\begin{IEEEbiography}[{\includegraphics[width=1in,height=1.25in,clip,keepaspectratio]{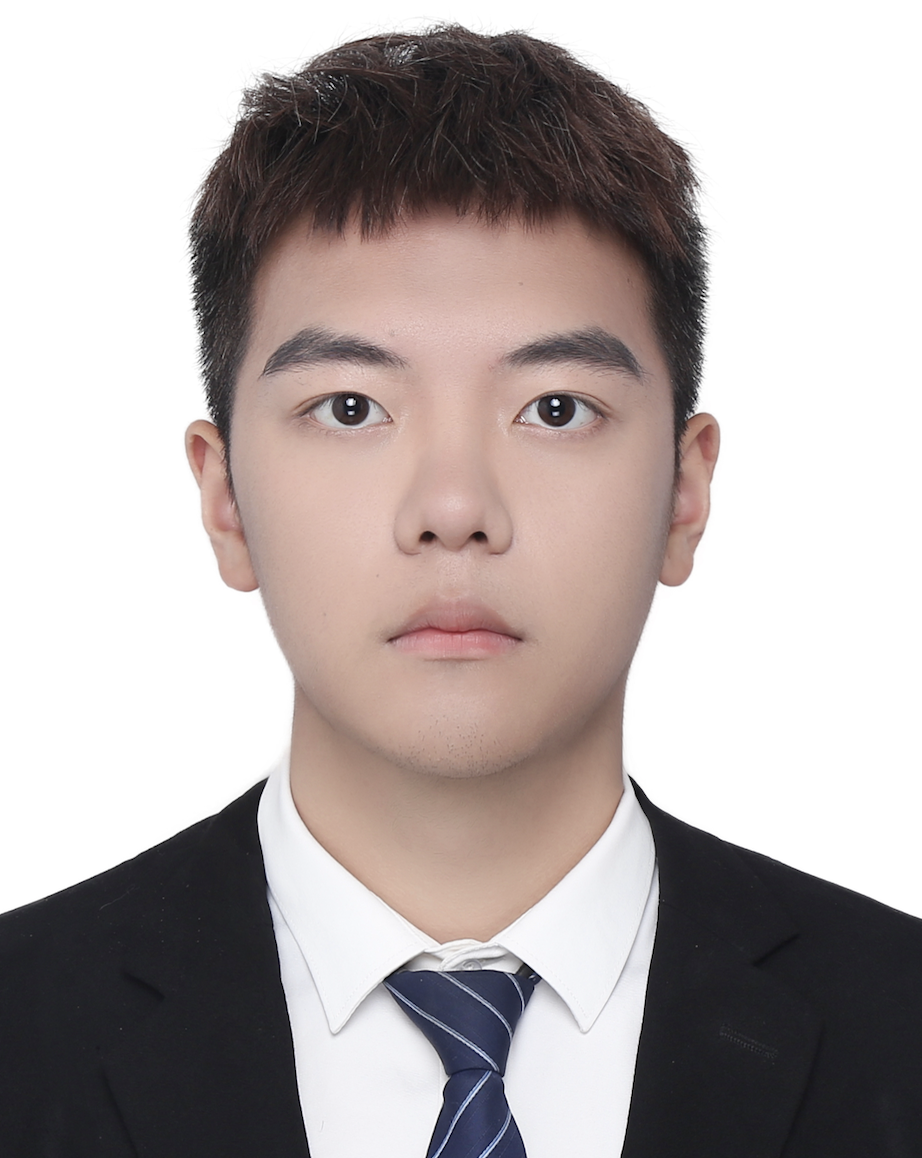}}]{Shuai Dong} received the B.Sc. degree (expected) in intelligent science and technology from China University of Geosciences, Wuhan, China. His research interests include computer vision, pattern recognition, machine learning, particularly include AI safety and trustworthiness.
\end{IEEEbiography}
\vspace{-22pt}

\begin{IEEEbiography}[{\includegraphics[width=1in,height=1.25in,clip,keepaspectratio]{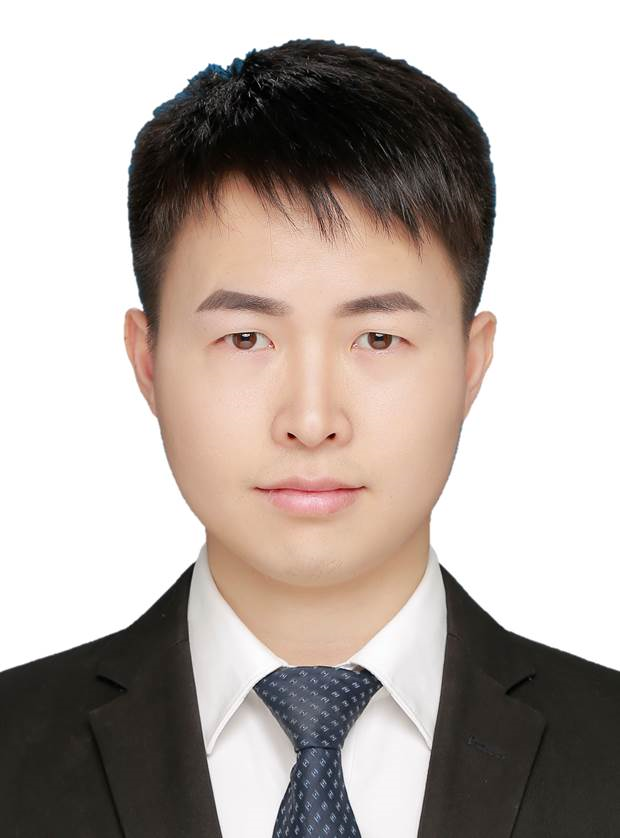}}]{Jie Zhang}
(Member, IEEE) received the Ph.D. degree from the University of Chinese Academy of Sciences (CAS), Beijing, China. He is currently an Associate Professor with the Institute of Computing Technology, CAS. His research interests include computer vision, pattern recognition, machine learning, particularly include adversarial attacks and defenses, domain generalization, AI safety and trustworthiness.
\end{IEEEbiography}
\vspace{-22pt}

\begin{IEEEbiography}[{\includegraphics[width=1in,height=1.25in,clip,keepaspectratio]{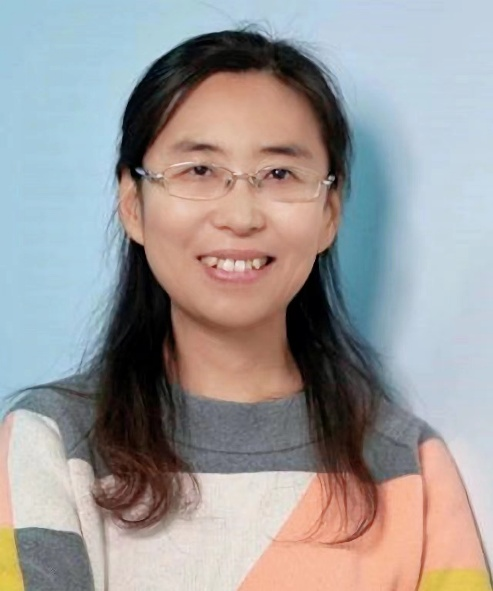}}]{Guoying Zhao} (Fellow, IEEE) received the Ph.D. degree in computer science from the Chinese Academy of Sciences, Beijing, China, in 2005. She is currently an Academy Professor and full Professor (tenured in 2017) with University of Oulu, and a PI with ELLIS Institute Finland. She is also a visiting professor with Aalto University. She is a member of Academia Europaea, a member of Finnish Academy of Sciences and Letters, Fellow of IEEE, IAPR, ELLIS and AAIA. She has authored or co-authored more than 360 papers in journals and conferences with 35490+ citations in Google Scholar and h-index 91. She is the recipient of 2024 IAPR Maria Petrou Prize. She has been associate Editor-in-Chief for Computer Vision and Image Understanding (CVIU), was/is associate editor for IEEE Trans. on Multimedia, Pattern Recognition, IEEE Trans. on Circuits and Systems for Video Technology, Image and Vision Computing and Frontiers in Psychology Journals. She is general co-chair for CVIP2025 and ACII 2025, was program co-chair for ACM International Conference on Multimodal Interaction (ICMI 2021), tutorial chair for ICPR 2024, panel chair for FG 2023, publicity chair of 22nd Scandinavian Conference on Image Analysis (SCIA 2023) and FG2018, and has served as area chairs for many conferences. Her current research interests include image and video representation, facial-expression and micro-expression recognition, emotional gesture analysis, affective computing, and biometrics. Her research has been reported by Finnish TV programs, newspapers and MIT Technology Review.
\end{IEEEbiography}
\vspace{-22pt}

\begin{IEEEbiography}[{\includegraphics[width=1in,height=1.25in,clip,keepaspectratio]{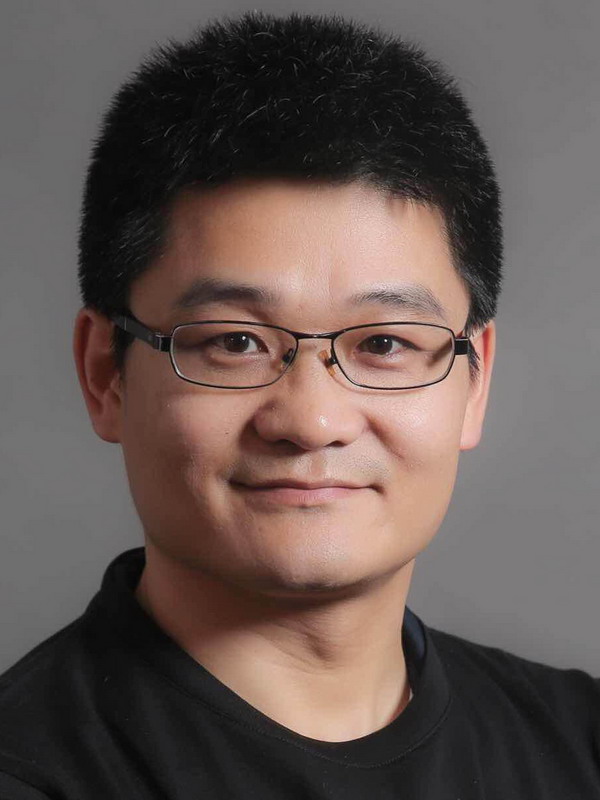}}]{Shiguang Shan}
(Fellow, IEEE) received the Ph.D. degree in computer science from the Institute of Computing Technology (ICT), Chinese Academy of Sciences (CAS), Beijing, China, in 2004. He has been a Full Professor with ICT since 2010, where he is currently the Director of the Key Laboratory of Intelligent Information Processing, CAS. His research interests include signal processing, computer vision, pattern recognition, and machine learning. He has published more than 300 articles in related areas. He served as the General Co-Chair for IEEE Face and Gesture Recognition 2023, the General Co-Chair for Asian Conference on Computer Vision (ACCV) 2022, and the Area Chair of many international conferences, including CVPR, ICCV, AAAI, IJCAI, ACCV, ICPR, and FG. He was/is an Associate Editors of several journals, including IEEE Transactions on Image Processing, Neurocomputing, CVIU, and PRL. He was a recipient of the China's State Natural Science Award in 2015 and the China’s State S\&T Progress Award in 2005 for his research work.
\end{IEEEbiography}
\vspace{-22pt}

\begin{IEEEbiography}[{\includegraphics[width=1in,height=1.25in,clip,keepaspectratio]{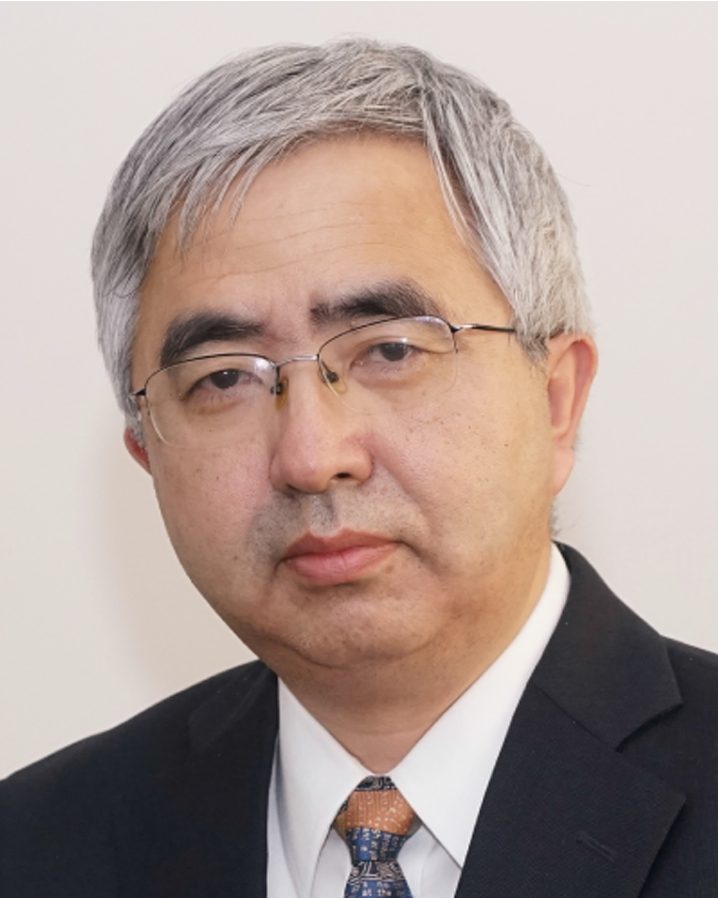}}]{Xilin Chen} (Fellow, IEEE) is currently a Professor with the Institute of Computing Technology, Chinese
 Academy of Sciences (CAS). He has authored one
 book and more than 400 articles in refereed journals
 and proceedings in the areas of computer vision,
 pattern recognition, image processing, and multi
modal interfaces. He is a fellow of the ACM,
 IAPR, and CCF. He is also an Information Sciences
 Editorial Board Member of Fundamental Research,
 an Editorial Board Member of Research, a Senior
 Editor of the Journal of Visual Communication and
 Image Representation, and an Associate Editor-in-Chief of the Chinese Jour
nal of Computers and Chinese Journal of Pattern Recognition and Artificial
 Intelligence. He served as an organizing committee member for multiple
 conferences, including the General Co-Chair of FG 2013/FG 2018, VCIP
 2022, the Program Co-Chair of ICMI 2010/FG 2024, and an Area Chair of
 ICCV/CVPR/ECCV/NeurIPS for more than ten times.
\end{IEEEbiography}



\vfill

\end{document}